%% file: cwavegan.tex
\documentclass{article}

\usepackage[square,numbers]{natbib}
\usepackage{float}
\usepackage[preprint]{nips_2018}

\usepackage[utf8]{inputenc} 
\usepackage[T1]{fontenc}    
\usepackage{hyperref}       
\usepackage{url}            
\usepackage{booktabs}       
\usepackage{amsfonts}       
\usepackage{nicefrac}       
\usepackage{microtype}      
\usepackage{graphicx}

\title{Conditional WaveGAN}

\author{
  Chae Young Lee \footnotemark[1]\\
  HAFS \\
  \texttt{cylee@hafs.hs.kr}
  \And
  Anoop Toffy  \thanks{These two authors contributed equally}\\
  IIIT Bangalore \\
  \texttt{anoop.toffy@iiitb.org} \\
  \And
  Gue Jun Jung \\
  SK Telecom \\
  \texttt{guejun.jung@sk.com}
  \And
  Woo-Jin Han \\
  Netmarble IGS \\
  \texttt{wjhan@igsinc.co.kr}
}

\begin{document}

\maketitle

\begin{abstract}
Generative models are successfully used for image synthesis \cite{huang2018introduction} in the recent years. But when it comes to other modalities like audio, text etc little progress has been made. Recent works focus on generating audio from a generative model in an unsupervised setting. We explore the possibility of using generative models conditioned on class labels. Concatenation based conditioning and conditional scaling were explored in this work with various hyper-parameter tuning methods \cite{salimans2016improved} \cite{kurach2018gan}. In this paper we introduce Conditional WaveGANs (cWaveGAN). Find our implementation at \url{https://github.com/acheketa/cwavegan}

\end{abstract}

\section{Introduction}

Generative adversarial networks \cite{NIPS2014_5423} are being widely used for synthesizing realistic images \cite{isola2017image, radford2015unsupervised, agustsson2018generative}. But very little has been explored in the area of audio generation. A few research works has been made in the area of unsupervised generative models in audio. One of them is WaveGAN \cite{donahue2018synthesizing}, which trains a generative model in an unsupervised setting.
In this work we use WaveGAN model as our baseline model. Audio samples generated from WaveGAN are human-recognizable and have a relatively good inception scores \cite{barratt2018note}. But the samples generated are completely random. 

The need for generating synthetic data has many applications in various fields. For example, in digital image enhancement, synthesizing new images can be used for smoothing (denoising), filling missing pieces (inpainting), improving resolution (super-resolution imaging) etc. Other areas include reinforcement learning, where agents can be trained with more training examples, and automatic speech recognition, where the synthesized data from generative models can be used as training data for training acoustic models. More direct usage is in the field of semi-supervised learning techniques \cite{schoneveld2017semi, odena2016semi, salimans2016improved}. In the recent 2018 ACM webinar \cite{IanGAN} Ian Goodfellow talks about the semi-supervised learning techniques and their applications in various fields.

In this work we explore a way to generate audio samples conditioned on class labels. That is, given a class label whether the generator of GAN can generate a particular audio waveform. In the history of GAN such type of conditioning has been explored in the synthesis of images. Conditional Generative Adversarial Nets introduced in 2016 by Mehdi Mirza et. al \cite{mirza2014conditional} was the first to try concatenation based conditioning. The conditional GANs were able to synthesize realistic images on MNIST and MIR Flickr datasets.

\section{Related Work}
Ever since the introduction of Generative Adversarial Networks by Ian Goodfellow et. al. \cite{NIPS2014_5423} in 2014 a lot of varieties of GANs has been implemented for accomplishing various tasks. It is one of the much explored generative models when compared to Variational Autoencoders (VAEs) and autoregressive models, assisted by a great tutorial given by Ian Goodfellow in NIPS 2016 \cite{goodfellow2016nips}. Most of the notable advancements were made in the field of image synthesis. In his 2018 paper, He Huang et al. \cite{huang2018introduction} talks more about the usage of GANs for image synthesis and various techniques in this fields.

Usage of generative models in natural language processing is also in progress with notabld fields including text modelling \cite{yu2017seqgan, rajeswar2017adversarial}, dialogue generation \cite{li2017adversarial}, and neural machine translation \cite{yang2017improving}. When it comes to audio generation, several approaches have been explored. Nonetheless, models that can generate audio waveforms directly (as opposed to some other representation
that can be converted into audio afterwards, such as spectrograms or piano rolls) are only starting to be explored. Autoregressive models were initially used for the generation of raw audio. WaveNet by Van Den Oord et al. \cite{van2016wavenet} is a convolutional model with dilated convolutions that learns to generate raw audio by autoregressive modelling. While a single WaveNet can capture the characteristics of many different speakers with equal fidelity and can switch between them by conditioning on the speaker identity, WaveRNN \cite{kalchbrenner2018efficient} describes a single-layer recurrent neural network with a dual softmax layer that matches the quality of the state-of-the-art WaveNet model. WaveNets have been applied to music generation as well. The paper by Sander Dieleman et al. (2018) \cite{dieleman2018challenge} talks about the challenges in realistic music generation.

The application of GAN in speech were used mainly for audio enhancements techniques and little on raw audio generation. SEGAN by Santiago Pascual et al \cite{pascual2017segan} uses deep networks which operates on waveform level, training the model end-to-end for denoising waveform chunks. The model works as a fully convolutional encoder-decoder structure. Tacotron \cite{wang2017tacotron} is an end-to-end generative text-to-speech model that synthesizes speech directly from characters. Since Tacotron generates speech at the frame level, it's substantially faster than sample-level autoregressive methods. In Tacotron2 \cite{shen2017natural} a neural network architecture synthesizes speech directly from the text. Tacotron2 achieves a mean opinion score (MOS) of 4.53, whereas a professionally recorded speech obtained a MOS of 4.58. 

In the recent work on speech synthesis, Chris Donahue et al. (2018) \cite{donahue2018synthesizing} introduces two GAN models, WaveGAN and SpecGAN. The waveGAN works in the time-domain and specGAN works in the frequency-domain. WaveGAN can produce intelligible words from a small vocabulary of human speech, as well as synthesize audio from other domains such as bird vocalizations, drums, and piano. It uses GANs in an unsupervised setting. WaveGAN is based on the DCGAN \cite{radford2015unsupervised} architecture, which became famous by its usage in image synthesis. We focus on this work to make the WaveGAN conditioned on class labels by introducing various conditioning techniques discussed in the following sections.

\section{Adversarial Nets} 

\subsection{Generative Adversarial Nets}

GAN was introduced as a novel method to train generative models. They are composed of two adversarial models: the \textit{generator, G} and the \textit{discriminator, D}. The generator, G captures the data distribution while the discriminator, D estimates the probability that the data generated by generator is coming from the data distribution or not. Both G and D can be non-linear mapping functions.

To learn the data distribution $P_{D}$ over data $x$, the generator builds a mapping from a noise distribution $P_{z}$ to data space as $G(z,\theta_{g})$. And the discriminator $D(x, \theta_{d})$ outputs as scalar representing that the training data $p_{x}$ rather than $p_{g}$.

${G}$ and ${D}$ are both trained simultaneously: we adjust parameters for ${G}$ to minimize ${log(1-D(G(\boldmath{z}))}$ and
adjust parameters for $D$ to minimize ${logD(X)}$, as if they are following the two-player min-max game with value function ${V(G, D)}$:

\begin{equation}
\label{eq:minimaxgame-definition}
\min_G \max_D V(D, G) = \mathbb{E}_{\boldmath{x} \sim p_{\textit{data}}(\boldmath{x})}[\log D(\boldmath{x})] + \mathbb{E}_{\boldmath{z} \sim p_z(\boldmath{z})}[\log (1 - D(G(\boldmath{z})))].
\end{equation}

\subsection{Conditional Generative Adversarial Nets}

The conditional GANs (cGANs)  \cite{mirza2014conditional} were introduced to make use of the additional information available in the form of labels, $y$. So instead of generating a random data from the generator, the auxiliary information like labels, $y$ are passed to both the generator and the discriminator along with the input features, ($z$ and $x$) so as to generate data conditioned on class labels. 

In both the generator and the discriminator, the label information, $y$ is augmented with the input features ($z$ and $x$).

The objective function of a two-player minimax game would be as Eq. \ref{eq:minimaxgame-definition-conditioned}. 

\begin{equation}
\label{eq:minimaxgame-definition-conditioned}
\min_G \max_D V(D, G) = \mathbb{E}_{\boldmath{x} \sim p_{\textit{data}}(\boldmath{x})}[\log D(\boldmath{x} | \boldmath{y})] + \mathbb{E}_{\boldmath{z} \sim p_z(\boldmath{z})}[\log (1 - D(G(\boldmath{z} | \boldmath{y})))].
\end{equation}

\section{Conditioning methodologies}

\begin{figure}[H]
  
  \centering
  \includegraphics[width=10cm]{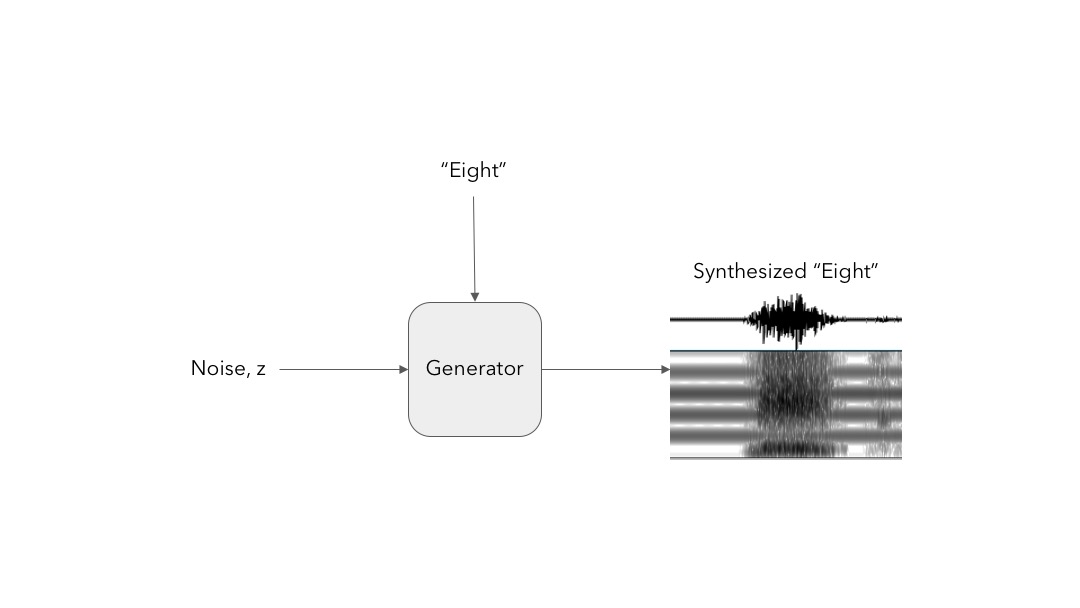}
  \caption{Synthesizing raw audio by conditioning}
  \label{figureTwo}
\end{figure}

In this work we explore a few conditioning mechanisms discussed in Dumoulin et. al \cite{dumoulin2018feature} article in this section. We are interested in generating raw audio of a particular given class label from the generator output. That is, the model takes as input a class and a source of random noise (e.g., a vector sampled from a normal distribution) and outputs a raw audio sample for the requested class as shown in Fig. \ref{figureTwo}.

\subsection{Concatenation based conditioning}

\begin{figure}[H]
  
  \centering
  \includegraphics[width=12cm]{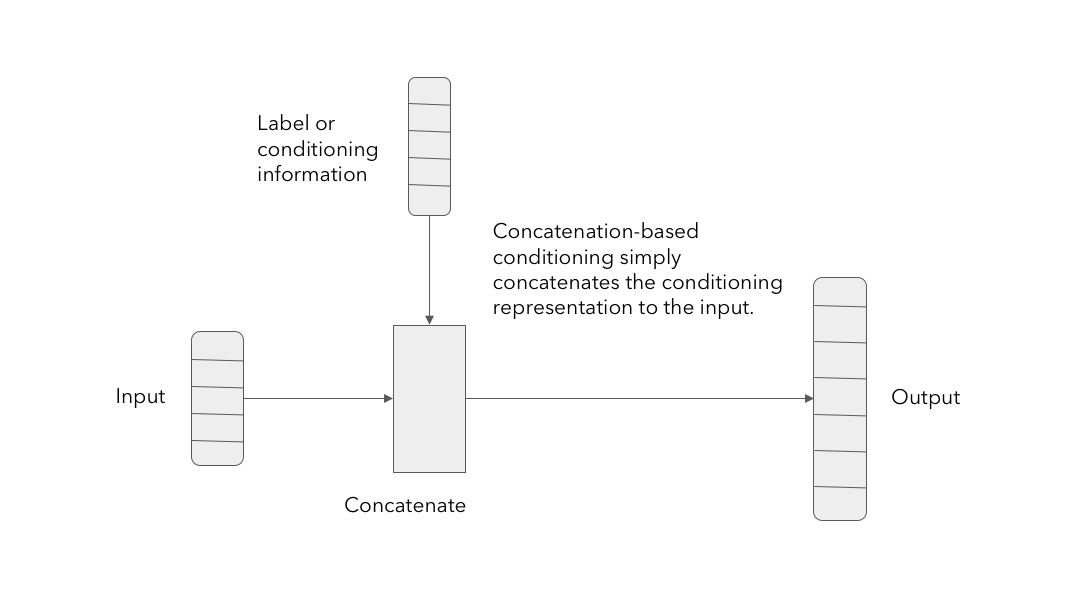}
  \caption{Concatenation based conditioning }
  \label{figureOne}
\end{figure}

First approach is to embed the class information to the input feature vector. That is, before training we would concatenate a representation of conditioning information to the noise vector and use it as the model's input as shown in Fig. \ref{figureOne}

\subsection{Conditional scaling}

In the other type of conditioning we implemented, we scaled the hidden layers based on the conditioning representation and multiplied it with the input vector for both the discriminator and the generator input. Also the scaling is applied to each layer of the convolution model.

\begin{figure}[H]
  \centering
  \includegraphics[width=12cm]{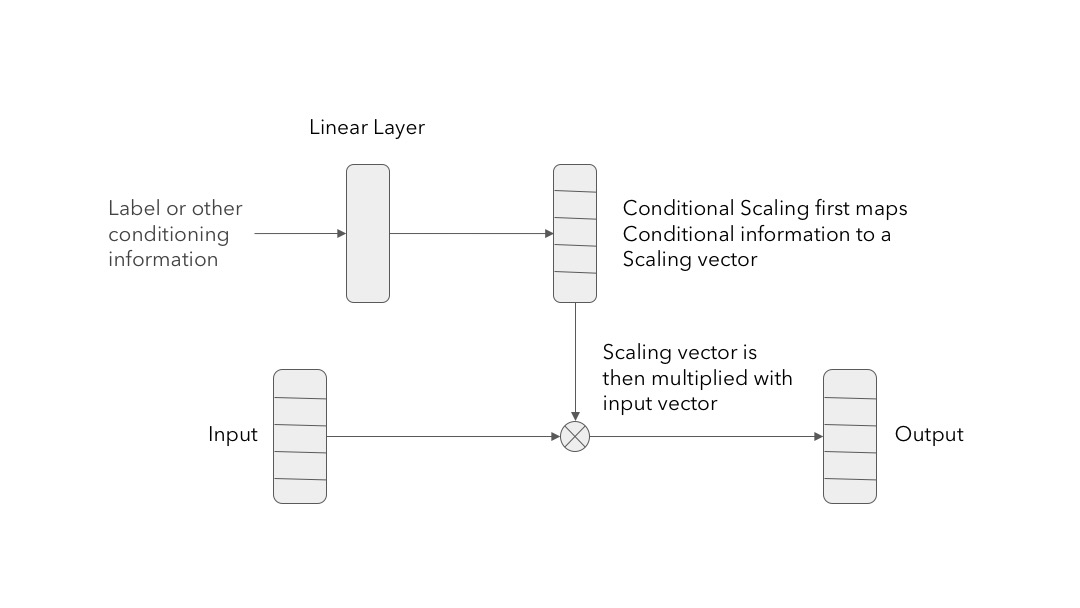}
  \caption{Conditional scaling}
  \label{figureThree}
\end{figure}

\section{Experimental Setting}

We based our architecture on WaveGAN model used by Chris Donahue et. al. \cite{donahue2018synthesizing} which inturn is based on the DCGAN model by Radford et. al \cite{radford2015unsupervised}. We modified the input vector dimension to 8192 by removing the beginning and ending silence to better capture the word alone. As in WaveGAN paper, we used one-dimensional filters of length 25 instead of two-dimensional filters of size 5x5 used in DCGAN model architecture but changed the strides from 4 to 2 just to incorporate the change in input vector dimension. The architecture of generator and discriminator model is provided in Table \ref{tab:garch} and Table \ref{tab:darch} respectively.

As a pilot study for this project we trained our model with DCGAN \cite{radford2015unsupervised} and WAGAN-GP \cite{gulrajani2017improved} losses with and without batch normalization. We compare two combinations in the following table \ref{tab:hyper}. We tried with different learning rate apart from $\alpha=2\mathrm{e}{-4}$, ie with $\alpha=2\mathrm{e}{-2}$ and $\alpha=4\mathrm{e}{-4}$. We also tried different learning rates with generator and disriminator since we found discriminator loss graph spiking.

We train our networks using batches of size 64 on four NVIDIA Telsa V100 accelerators and batches of 1024 on a single 8 core TPUv2. We trained it over 200 epochs for around 3 days.

\begin{table}[t]
\caption{Conditional WaveGAN generator architecture}
\centering
\footnotesize
\begin{tabular}{ l | l | l }
Operation & Kernel Size & Output Shape \\
\hline
\input{garch.tex}
\end{tabular}
\label{tab:garch}
\end{table}

\begin{table}[t]
\caption{Conditional WaveGAN discriminator architecture}
\centering
\footnotesize
\begin{tabular}{ l | l | l }
Operation & Kernel Size & Output Shape \\
\hline
\input{darch.tex}
\end{tabular}
\label{tab:darch}
\end{table}

\begin{table}[t!]
\caption{Conditional WaveGAN hyperparameters}
\centering
\footnotesize
\begin{tabular}{ l | l | l | l | l}
Name & Value (GPU) &  Value (GPU)  & Value (TPU) & Value (TPU)\\
\hline
Input data type & $16$-bit PCM & $16$-bit PCM & $16$-bit PCM & $16$-bit PCM \\
Model data type & $32$-bit float &  $32$-bit float  &  $32$-bit float  &  $32$-bit float \\
Num channels ($c$) & 1 & 1 & 1 & 1 \\
Batch size ($b$) & 64 & 64 & 1024 & 1024\\
Model size ($d$) & 64 & 64 & 64 & 64\\
Phase shuffle (WaveGAN) & 2 & 2 & 2 & 2 \\
Loss & WGAN-GP~\citep{gulrajani2017improved} & DCGAN \cite{radford2015unsupervised} & WGAP-GP & DCGAN \\
$D$ updates per $G$ update & 5  & 5 & 5 & 5\\
Optimizer & Adam ($\alpha=1\mathrm{e}{-4}$) & Adam ($\alpha=2\mathrm{e}{-4}$) & Adam ($\alpha=2\mathrm{e}{-4}$) & Adam ($\alpha=2\mathrm{e}{-4}$)\\
\end{tabular}
\label{tab:hyper}
\end{table}

\section{Dataset}
We used the \textit{Speech Commands
Dataset} \cite{buchner2017parametric} released by Google AI Team for conducting our priliminary experiments. This dataset consists of many speakers recording individual words in uncontrolled recording conditions. We used a subset of the dataset as done by the authors of WaveGAN paper, ie. Speech Commands Zero Through Nine (SC09) subset, which reduces the vocabulary of the dataset to ten words: the digits “zero” through “nine”.

Each recording utterance is of one second in length and the training set consists of 1850 recording for each word adding upto 5.3hr of speech data. The wide variety of alignments, speakers and recording conditions make this a challenging dataset from a generative perspective.

\section{Preliminary Results}
We were able to generate human realistic raw audio from the conditional WaveGAN. But most of the time we faced the difficulty of having generated noisy output. The conditioning that we tried was not consistent more often and we ended up generating a false positive sample. So we will research further to stabilize the GAN training and to tune our training parameters to generate raw audio out that are realistic to real world data. 

This is a pilot study which explores the possibility of making use of synthesized data from generative models as training data by other discriminative models.

\section{Conclusion}
Synthesizing audio is a area that has been growing recently and is much focused by researcher since it has its application in building robust speech recognition systems and improved text-to-speech systems. Much research were focused on generation on random audio from unlabeled training data. In this work we focused on generation of audio given a class label. We were able to get human recognizable synthesized audio waveform from conditional waveGANs, but the presence of noise in the sample generated needs to be smoothed as a continuation of the present work, which gives a lot of scope for future improvements. Secondly, hyperparameter tuning of GAN needs to be explored since presently the training is unstable. Thirdly, the conditioning method explored so far is increasing training time of GANs, so a more novel conditioning method need to be discovered.

We hope this work gives direction to more research on conditional generation of audio wave-forms and gets extended to other frontiers.

\subsubsection*{Acknowledgments}

We would like thank Tensorflow Korea and Google Korea, and all other sponsors of Deep Learning Camp Jeju 2018,  \url{http://jeju.dlcamp.org/2018/}. This work was done as a part of the DL Jeju Camp. We would like to convey our sincere gratitude to our mentors, Dr.Gue Jun Jung (SK Telecom), Dr. Woo-Jin Han (Netmarble IGS) and all other men-tees of the camp. 

We are also delighted to use TPUs \cite{tpus} for our training process which made a considerable improvement in making the training process faster.


\end{document}

%% file: garch.tex
Input $\boldmath{z} \sim \textit{Uniform}(-1, 1)$ &  & ($n$, $100$)\\
Dense 1 & ($100$, $256d$) & ($n$, $256d$)\\
Reshape & & ($n$, $16$, $16d$)\\
ReLU & & ($n$, $16$, $16d$)\\
Trans Conv1D (Stride=$4$) & ($25$, $16d$, $8d$) & ($n$, $64$, $8d$)\\
ReLU & & ($n$, $64$, $8d$)\\
Trans Conv1D (Stride=$4$) & ($25$, $8d$, $4d$) & ($n$, $256$, $4d$)\\
ReLU & & ($n$, $256$, $4d$)\\
Trans Conv1D (Stride=$4$) & ($25$, $4d$, $2d$) & ($n$, $1024$, $2d$)\\
ReLU & & ($n$, $1024$, $2d$)\\
Trans Conv1D (Stride=$4$) & ($25$, $2d$, $d$) & ($n$, $4096$, $d$)\\
ReLU & & ($n$, $4096$, $d$)\\
Trans Conv1D (Stride=$2$) & ($25$, $d$, $c$) & ($n$, $8192$, $c$)\\
Tanh & & ($n$, $8192$, $c$)


%% file: darch.tex
Input $\boldmath{x}$ or $G(\boldmath{z})$ & & ($n$, $8192$, $c$)\\
Conv1D (Stride=$2$) & ($25$, $c$, $d$) & ($n$, $4096$, $d$) \\
LReLU ($\alpha = 0.2$) & & ($n$, $4096$, $d$) \\
Phase Shuffle ($n = 2$) & & ($n$, $4096$, $d$) \\
Conv1D (Stride=$4$) & ($25$, $d$, $2d$) & ($n$, $1024$, $2d$) \\
LReLU ($\alpha = 0.2$) & & ($n$, $1024$, $2d$) \\
Phase Shuffle ($n = 2$) & & ($n$, $1024$, $2d$) \\
Conv1D (Stride=$4$) & ($25$, $2d$, $4d$) & ($n$, $256$, $4d$) \\
LReLU ($\alpha = 0.2$) & & ($n$, $256$, $4d$) \\
Phase Shuffle ($n = 2$) & & ($n$, $256$, $4d$) \\
Conv1D (Stride=$4$) & ($25$, $4d$, $8d$) & ($n$, $64$, $8d$) \\
LReLU ($\alpha = 0.2$) & & ($n$, $64$, $8d$) \\
Phase Shuffle ($n = 2$) & & ($n$, $64$, $8d$) \\
Conv1D (Stride=$4$) & ($25$, $8d$, $16d$) & ($n$, $16$, $16d$) \\
LReLU ($\alpha = 0.2$) & & ($n$, $16$, $16d$) \\
Reshape & & ($n$, $256d$) \\
Dense & ($256d$, $1$) & ($n$, $1$) \\

%% file: cwavegan.bbl
\begin{thebibliography}{28}
\providecommand{\natexlab}[1]{#1}
\providecommand{\url}[1]{\texttt{#1}}
\expandafter\ifx\csname urlstyle\endcsname\relax
  \providecommand{\doi}[1]{doi: #1}\else
  \providecommand{\doi}{doi: \begingroup \urlstyle{rm}\Url}\fi

\bibitem[Agustsson et~al.(2018)Agustsson, Tschannen, Mentzer, Timofte, and
  Van~Gool]{agustsson2018generative}
E.~Agustsson, M.~Tschannen, F.~Mentzer, R.~Timofte, and L.~Van~Gool.
\newblock Generative adversarial networks for extreme learned image
  compression.
\newblock \emph{arXiv preprint arXiv:1804.02958}, 2018.

\bibitem[Barratt and Sharma(2018)]{barratt2018note}
S.~Barratt and R.~Sharma.
\newblock A note on the inception score.
\newblock \emph{arXiv preprint arXiv:1801.01973}, 2018.

\bibitem[Buchner(2017)]{buchner2017parametric}
J.~Buchner.
\newblock Synthetic speech commands dataset.
\newblock
  \emph{https://www.kaggle.com/jbuchner/synthetic-speech-commands-dataset},
  2017.

\bibitem[Dieleman et~al.(2018)Dieleman, Oord, and
  Simonyan]{dieleman2018challenge}
S.~Dieleman, A.~v.~d. Oord, and K.~Simonyan.
\newblock The challenge of realistic music generation: modelling raw audio at
  scale.
\newblock \emph{arXiv preprint arXiv:1806.10474}, 2018.

\bibitem[Donahue et~al.(2018)Donahue, McAuley, and
  Puckette]{donahue2018synthesizing}
C.~Donahue, J.~McAuley, and M.~Puckette.
\newblock Synthesizing audio with generative adversarial networks.
\newblock \emph{arXiv preprint arXiv:1802.04208}, 2018.

\bibitem[Dumoulin et~al.(2018)Dumoulin, Perez, Schucher, Strub, Vries,
  Courville, and Bengio]{dumoulin2018feature}
V.~Dumoulin, E.~Perez, N.~Schucher, F.~Strub, H.~d. Vries, A.~Courville, and
  Y.~Bengio.
\newblock Feature-wise transformations.
\newblock \emph{Distill}, 3\penalty0 (7):\penalty0 e11, 2018.

\bibitem[Goodfellow(2016)]{goodfellow2016nips}
I.~Goodfellow.
\newblock Nips 2016 tutorial: Generative adversarial networks.
\newblock \emph{arXiv preprint arXiv:1701.00160}, 2016.

\bibitem[Goodfellow et~al.(2014)Goodfellow, Pouget-Abadie, Mirza, Xu,
  Warde-Farley, Ozair, Courville, and Bengio]{NIPS2014_5423}
I.~Goodfellow, J.~Pouget-Abadie, M.~Mirza, B.~Xu, D.~Warde-Farley, S.~Ozair,
  A.~Courville, and Y.~Bengio.
\newblock Generative adversarial nets.
\newblock In Z.~Ghahramani, M.~Welling, C.~Cortes, N.~D. Lawrence, and K.~Q.
  Weinberger, editors, \emph{Advances in Neural Information Processing Systems
  27}, pages 2672--2680. Curran Associates, Inc., 2014.
\newblock URL
  \url{http://papers.nips.cc/paper/5423-generative-adversarial-nets.pdf}.

\bibitem[Google(2016)]{tpus}
Google.
\newblock Cloud tpu.
\newblock \emph{https://cloud.google.com/tpu/}, 2016.

\bibitem[Gulrajani et~al.(2017)Gulrajani, Ahmed, Arjovsky, Dumoulin, and
  Courville]{gulrajani2017improved}
I.~Gulrajani, F.~Ahmed, M.~Arjovsky, V.~Dumoulin, and A.~C. Courville.
\newblock Improved training of wasserstein gans.
\newblock In \emph{Advances in Neural Information Processing Systems}, pages
  5767--5777, 2017.

\bibitem[Huang et~al.(2018)Huang, Yu, and Wang]{huang2018introduction}
H.~Huang, P.~S. Yu, and C.~Wang.
\newblock An introduction to image synthesis with generative adversarial nets.
\newblock \emph{arXiv preprint arXiv:1803.04469}, 2018.

\bibitem[Ian~Goodfellow(2016)]{IanGAN}
G.~B. Ian~Goodfellow, Staff Research~Scientist.
\newblock Adversarial machine learning.
\newblock \emph{https://goo.gl/NVWF7j}, 2016.

\bibitem[Isola et~al.(2017)Isola, Zhu, Zhou, and Efros]{isola2017image}
P.~Isola, J.-Y. Zhu, T.~Zhou, and A.~A. Efros.
\newblock Image-to-image translation with conditional adversarial networks.
\newblock \emph{arXiv preprint}, 2017.

\bibitem[Kalchbrenner et~al.(2018)Kalchbrenner, Elsen, Simonyan, Noury,
  Casagrande, Lockhart, Stimberg, Oord, Dieleman, and
  Kavukcuoglu]{kalchbrenner2018efficient}
N.~Kalchbrenner, E.~Elsen, K.~Simonyan, S.~Noury, N.~Casagrande, E.~Lockhart,
  F.~Stimberg, A.~v.~d. Oord, S.~Dieleman, and K.~Kavukcuoglu.
\newblock Efficient neural audio synthesis.
\newblock \emph{arXiv preprint arXiv:1802.08435}, 2018.

\bibitem[Kurach et~al.(2018)Kurach, Lucic, Zhai, Michalski, and
  Gelly]{kurach2018gan}
K.~Kurach, M.~Lucic, X.~Zhai, M.~Michalski, and S.~Gelly.
\newblock The gan landscape: Losses, architectures, regularization, and
  normalization.
\newblock \emph{arXiv preprint arXiv:1807.04720}, 2018.

\bibitem[Li et~al.(2017)Li, Monroe, Shi, Jean, Ritter, and
  Jurafsky]{li2017adversarial}
J.~Li, W.~Monroe, T.~Shi, S.~Jean, A.~Ritter, and D.~Jurafsky.
\newblock Adversarial learning for neural dialogue generation.
\newblock \emph{arXiv preprint arXiv:1701.06547}, 2017.

\bibitem[Mirza and Osindero(2014)]{mirza2014conditional}
M.~Mirza and S.~Osindero.
\newblock Conditional generative adversarial nets.
\newblock \emph{arXiv preprint arXiv:1411.1784}, 2014.

\bibitem[Odena(2016)]{odena2016semi}
A.~Odena.
\newblock Semi-supervised learning with generative adversarial networks.
\newblock \emph{arXiv preprint arXiv:1606.01583}, 2016.

\bibitem[Pascual et~al.(2017)Pascual, Bonafonte, and Serra]{pascual2017segan}
S.~Pascual, A.~Bonafonte, and J.~Serra.
\newblock Segan: Speech enhancement generative adversarial network.
\newblock \emph{arXiv preprint arXiv:1703.09452}, 2017.

\bibitem[Radford et~al.(2015)Radford, Metz, and
  Chintala]{radford2015unsupervised}
A.~Radford, L.~Metz, and S.~Chintala.
\newblock Unsupervised representation learning with deep convolutional
  generative adversarial networks.
\newblock \emph{arXiv preprint arXiv:1511.06434}, 2015.

\bibitem[Rajeswar et~al.(2017)Rajeswar, Subramanian, Dutil, Pal, and
  Courville]{rajeswar2017adversarial}
S.~Rajeswar, S.~Subramanian, F.~Dutil, C.~Pal, and A.~Courville.
\newblock Adversarial generation of natural language.
\newblock \emph{arXiv preprint arXiv:1705.10929}, 2017.

\bibitem[Salimans et~al.(2016)Salimans, Goodfellow, Zaremba, Cheung, Radford,
  and Chen]{salimans2016improved}
T.~Salimans, I.~Goodfellow, W.~Zaremba, V.~Cheung, A.~Radford, and X.~Chen.
\newblock Improved techniques for training gans.
\newblock In \emph{Advances in Neural Information Processing Systems}, pages
  2234--2242, 2016.

\bibitem[Schoneveld(2017)]{schoneveld2017semi}
L.~Schoneveld.
\newblock Semi-supervised learning with generative adversarial networks.
\newblock 2017.

\bibitem[Shen et~al.(2017)Shen, Pang, Weiss, Schuster, Jaitly, Yang, Chen,
  Zhang, Wang, Skerry-Ryan, et~al.]{shen2017natural}
J.~Shen, R.~Pang, R.~J. Weiss, M.~Schuster, N.~Jaitly, Z.~Yang, Z.~Chen,
  Y.~Zhang, Y.~Wang, R.~Skerry-Ryan, et~al.
\newblock Natural tts synthesis by conditioning wavenet on mel spectrogram
  predictions.
\newblock \emph{arXiv preprint arXiv:1712.05884}, 2017.

\bibitem[Van Den~Oord et~al.(2016)Van Den~Oord, Dieleman, Zen, Simonyan,
  Vinyals, Graves, Kalchbrenner, Senior, and Kavukcuoglu]{van2016wavenet}
A.~Van Den~Oord, S.~Dieleman, H.~Zen, K.~Simonyan, O.~Vinyals, A.~Graves,
  N.~Kalchbrenner, A.~W. Senior, and K.~Kavukcuoglu.
\newblock Wavenet: A generative model for raw audio.
\newblock In \emph{SSW}, page 125, 2016.

\bibitem[Wang et~al.(2017)Wang, Skerry-Ryan, Stanton, Wu, Weiss, Jaitly, Yang,
  Xiao, Chen, Bengio, et~al.]{wang2017tacotron}
Y.~Wang, R.~Skerry-Ryan, D.~Stanton, Y.~Wu, R.~J. Weiss, N.~Jaitly, Z.~Yang,
  Y.~Xiao, Z.~Chen, S.~Bengio, et~al.
\newblock Tacotron: Towards end-to-end speech synthesis.
\newblock \emph{arXiv preprint arXiv:1703.10135}, 2017.

\bibitem[Yang et~al.(2017)Yang, Chen, Wang, and Xu]{yang2017improving}
Z.~Yang, W.~Chen, F.~Wang, and B.~Xu.
\newblock Improving neural machine translation with conditional sequence
  generative adversarial nets.
\newblock \emph{arXiv preprint arXiv:1703.04887}, 2017.

\bibitem[Yu et~al.(2017)Yu, Zhang, Wang, and Yu]{yu2017seqgan}
L.~Yu, W.~Zhang, J.~Wang, and Y.~Yu.
\newblock Seqgan: Sequence generative adversarial nets with policy gradient.
\newblock In \emph{AAAI}, pages 2852--2858, 2017.

\end{thebibliography}
